\documentclass[a4paper,twoside]{article}

\usepackage{lipsum}
\usepackage{epsfig}
\usepackage{subcaption}
\usepackage{calc}
\usepackage{amssymb}
\usepackage{amstext}
\usepackage{amsmath}
\usepackage{amsthm}
\usepackage{multicol}
\usepackage{pslatex}
\usepackage{apalike}
\usepackage[bottom]{footmisc}
\usepackage[table,xcdraw]{xcolor}
\usepackage{makecell}
\usepackage{hyperref}
\usepackage{cite}
\usepackage{float}
\usepackage{SCITEPRESS}     
\begin{document}

\title{Creation and Evaluation of a Food Product Image Dataset\newline for Product Property Extraction}

\author{\authorname{Christoph Brosch, Alexander Bouwens, Sebastian Bast, Swen Haab and Rolf Krieger}
\affiliation{Institut für Softwaresysteme, Hochschule Trier, Standort Birkenfeld \newline55768 Hoppstädten-Weiersbach, Germany}
\email{\{c.brosch, s19cc8, s.bast, s.haab, r.krieger\}@umwelt-campus.de}
}

\abstract{The enormous progress in the field of artificial intelligence (AI) enables retail companies to automate their processes and thus to save costs. Thereby, many AI-based automation approaches are based on machine learning and computer vision. The realization of such approaches requires high-quality training data. In this paper, we describe the creation process of an annotated dataset that contains 1,034 images of single food products, taken under studio conditions, annotated with 5 class labels and 30 object detection labels, which can be used for product recognition and classification tasks. We based all images and labels on standards presented by GS1, a global non-profit organisation. The objective of our work is to support the development of machine learning models in the retail domain and to provide a reference process for creating the necessary training data.}

\onecolumn \maketitle \normalsize \setcounter{footnote}{0} \vfill

\section{\uppercase{Introduction}}
\label{sec:Introduction}

The retail sector faces numerous challenges and opportunities due to rapid digitization and technological advancements. The growth of e-commerce has significantly impacted traditional retail, while advances in artificial intelligence (AI) offer retailers the opportunity to automate various processes. Assortment planning, pricing, and promotion planning as well as in-store logistics operations are just a few areas where AI can be applied.

To address these challenges and capitalize on new opportunities, retail companies are increasingly exploring automation concepts for their stores. Many of these solutions utilize computer vision and machine learning techniques for tasks such as product detection and recognition. Applications range from identifying missing products that need restocking to ensuring planogram compliance.

As a result, the amount of data to be managed per product has grown substantially, placing high requirements on product information management and master data management. Product images, in particular, have become increasingly important and must be effectively managed by product information management teams. They must guarantee that product data stored in systems are consistent with the data shown on the image.

To address this issue, one approach is to determine the product's properties from its image automatically. These properties include the product's name, brand, nutrition facts table, filling quantity, and category. The processing involves detecting and recognizing the product in the image, identifying image regions that describe the product properties, and extracting relevant information through various approaches based on machine learning.

A system that solves this problem can support numerous processes within a retail company, such as generating structured data describing a product based on its image, reducing manual data entry efforts in ERP, PIM, and online shop systems. Additionally, the extracted data can be used to verify whether system data matches the assigned product image, thus avoiding the display of incorrect or outdated images in online stores. The training of such models requires a large amount of training data.

In this paper, we give a short overview of several existing datasets in section 2 and explain why they are not suitable for our use case. Afterwards, in section 3, we present our product selection process in detail, providing researchers with a guideline on how to extend this dataset to cater to their economic or scientific needs. Section 4 describes the annotation process, while section 5 presents the dataset's statistical facts. In section 6, we introduce our baseline models trained on the dataset.

Finally, we discuss the workload predicted from our experiences to extend this dataset, as well as outline potential future work and conclude this paper.

\section{\uppercase{Related Work}}
\label{sec:RelatedWork}

In recent years, the research community has made notable advancements in the creation of datasets for product detection and recognition. These datasets are based on images featuring products in densely packed scenes, such as retail shelves or grouped items on a table for product detection \cite{goldman2019dense, follmann_mvtec_nodate} or a combination of both densely packed scenes as well as single product images \cite{4270484, george_recognizing_2014, georgiadis_products-6k_2021, wei_rpc_2019}. Product recognition entails assigning one or more classes from a non-fine-grained classification scheme to the detected product, or using its visual feature embedding to determine the exact product. Recent work by Chen et al. \cite{chen_unitail_2022} highlights the relevance of product detection and recognition in real-world shelf scenarios. The authors present an end-to-end system for detecting products on shelves and subsequently recognizing them based on the text extracted from the cropped product images. In contrast to this, we focus on single product images in our work. Next, we provide an overview of common datasets by highlighting their primary features, focusing on datasets containing single product images exclusively or in combination with images depicting densely packed scenes in a realistic environment. Afterwards, we distinguish our datasets from them. A recent survey\cite{wei_deep_2020} observed current problems and trends in the field of deep learning for product recognition and discusses existing datasets in more detail than we do here. The following list is a subset of the datasets described in \cite{wei_deep_2020}:

\begin{itemize}
    \item The Grozi-120 dataset \cite{4270484} consists of 120 products, each associated with multiple reference images as well as one or more videos, separated into several frames.
    \item The Grozi-3.2k dataset \cite{george_recognizing_2014} provides 3,235 images depicting Swiss retail shelves and $8,350$ images of reference products, showing the product only from its front face. The authors also provide detailed product detection annotations, for each of the shelf images. 
    \item The Retail Product Checkout dataset (RPC) \cite{wei_rpc_2019}, comprises $53,739$ single product images, depicting Chinese products, showing the product from four different vertical angles, while rotating from $0^\circ$ to $360^\circ$ on a rotary plate, resulting in 160 images for each of the $200$ unique products. The dataset also includes $30,000$ checkout images showing products in various constellations on a white plane.
    \item The MVTec D2S dataset \cite{follmann_mvtec_nodate} consists of $21,000$ images, displaying one or more products from one or more categories, depending on the associated set (train, validation or test), placed on a rotary plate with the picture taken from straight above.
\end{itemize}
We want to highlight two additional relevant datasets released in recent years:
\begin{itemize}
    \item The Products-6k dataset \cite{georgiadis_products-6k_2021} provides $2,917$ single-product images, depicting Greece products, each associated with one or more images presenting the product from one or more angles. It also contains $373$ query images showing products held by the photographer in front of grocery store floors or shelves.
    \item The ABO dataset \cite{collins2022abo} provided by Amazon, which annotates around 400,000 3D objects from 567 different classes. The 3D Objects depict a wide range of products and are not exclusive to grocery products.
\end{itemize}

Despite existing datasets like Products-6k\cite{georgiadis_products-6k_2021} and Retail Product Checkout\cite{wei_rpc_2019} offering various angles and high-resolution images, they have the disadvantage that they lack English or German text, professional backgrounds and lighting. Therefore, these datasets do not meet our research needs for e.g. performing an accurate optical character recognition (OCR). We require high-quality, individual product images from different angles with detailed annotations for regions of interest. To address this gap, we developed a dataset with optimal lighting, neutral backgrounds, and annotations for object recognition, image classification, and ground truth text values to evaluate end-to-end system performance.

\section{\uppercase{Image Collection}}
\label{sec:ImageCollection}
In the following subsections, we describe the procedure for selecting product categories and products that should be included in our dataset. Moreover, we describe how product images were captured and explain which images were created for each product. 
Mainly, our selection procedure is based on the standards provided by GS1. We took into account the Global Product Classification (GPC) standard \cite{noauthor_gpc_nodate}, the GS1 specification for product image creation \cite{noauthor_gs1_nodate}, the GS1 codes for packaging types \cite{netherlands_codes_nodate}, and the GS1 web vocabulary for food, beverage, and tobacco products \cite{noauthor_gs1foodbeveragetobaccoproduct_nodate}. Due to this fact, we assume that our dataset can be easily extended and used by other researchers.

\subsection{Product Category Selection}
\label{sec:ImageCollection:subsec:ProductCategorySelection}
The selection of product categories is based on the four-level hierarchical Global Product Classification system version 11/2020 \cite{noauthor_gpc_nodate}. The system corresponds to a mono-hierarchy. The categories of the four levels are called segments, families, classes, and bricks, respectively. Food products belong to the segment Food/Beverage/Tobacco. Table \ref{tab:gpc hierarchy} shows the number of categories per level. 
In our dataset, we focused on selecting products belonging to each family within the segment Food/Beverage/Tobacco, as products from this segment share properties which products of other segments do not have, such as nutrition tables or ingredient lists. In contrast, products from this segment usually do not have properties that products of other segments have, such as energy labels or hazard labels. Moreover, we omitted two families categorized within this segment from our dataset. 
\begin{itemize}
    \item Fresh Garnish (Food): Example bricks within this segment are "Banana Leaves" or "Orange Blossom", which are not commonly sold in grocery stores in our area. These circumstances made it difficult to acquire enough products from local grocery stores that fit in this GPC family.
    \item Tobacco/Cannabis/Smoking Accessories: Packaging for tobacco products often have different properties than packaging of food and beverages, such as hazard labels, warning labels, and graphic anti-smoking images, as mandated by European law. This meant that the labels found on cigarette packaging did not match those of the other products in the dataset.
\end{itemize}

\begin{table*}[t]
    \vspace{-0.2cm}
    
    \centering
    \caption{Structure of the Global Product Classification (GPC) standard version 11/2020. The third column shows the number of families, classes and bricks associated with segment Food/Beverage/Tobacco. The fourth column highlights the coverage of those classes in this dataset.}
    \begin{tabular}{|l|l|l|l|l|l|} \hline
    hierarchy level & \#classes & \#food & coverage & description & example                   \\ \hline
    segment         & 40        & 1     & 1                      & industry sector              & Food/Beverage/Tobacco     \\ \hline
    family          & 145       & 23    & 19                     & division of a segment        & Beverages                 \\ \hline
    class           & 908       & 131   & 62                    & group of similar categories  & Coffee/Coffee Substitutes \\ \hline
    brick           & 5,039      & 826  & 100                    & category of similar products & Coffee - Ground Beans \\ \hline
    \end{tabular}
    \label{tab:gpc hierarchy} 
\end{table*}

\subsection{Product Selection}
\label{sec:ImageCollection:subsec:ProductSelection}

We selected the products for our dataset following different criteria. Firstly, we aimed for each family within segment Food/Beverage/Tobacco to be represented by at least 10 products. An exception to this rule was made for families containing nuts, fruits, and vegetables. Within all three of these food categories, GPC offers different families labelled "Unprepared/Unprocessed (Fresh)" and "Unprepared/Unprocessed (Shelf Stable)". The only perceivable difference in these families is that the families labelled "Fresh" contain more specific bricks, whereas the families labelled "Shelf Stable" usually contain a single, more generic brick. In these cases, we decided to pick a total of 10 products for each category. When assigning the products to a brick, we always chose the most fitting option, only opting for the more generic brick, when none of the specific bricks fit the product. This means that the goal of 10 products per family has not yet been reached within these families.

The products were taken from 4 different German retail stores, in order to have products from different name and store brands, as well as different packaging types represented in our dataset.
 
\subsection{Product Images}
\label{sec:ImageCollection:subsec:ProductImages}
\begin{figure}
    \centering
    \includegraphics[width=\linewidth]{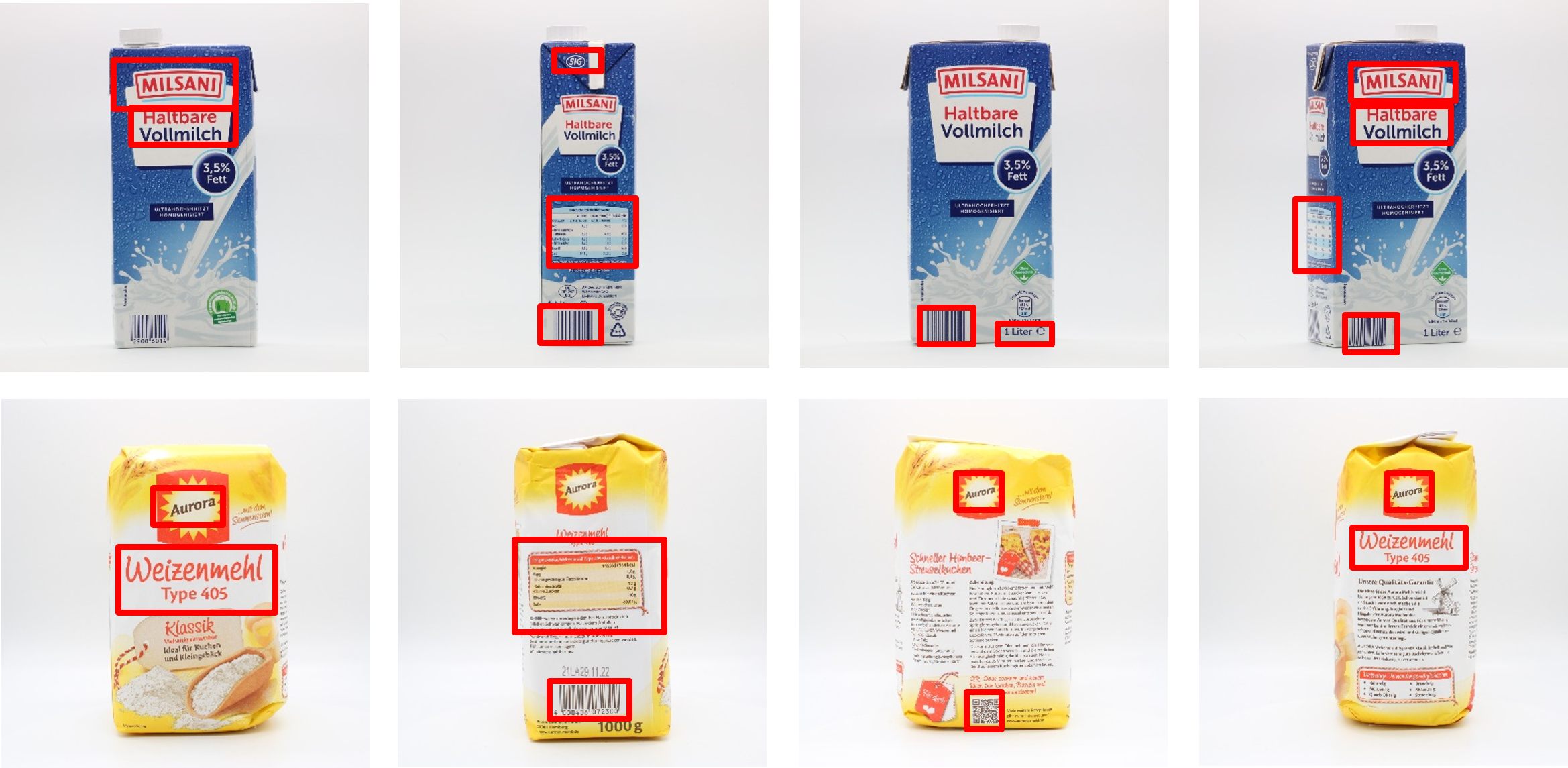}
    \caption{Example images from our dataset with visualized bounding boxes.}
    \label{fig:ExampleImagesDataset}
\end{figure}

In creating our images, we followed the GS1 product image specification standard for creating product images \cite{noauthor_gs1_nodate}. It defines, how to create and name product images under optimal conditions. We used this part of the specification to determine, how to create product images, however, we decided to use our own naming convention for product images. The specification also details different types of product images, as well as guidelines for image resolution, image quality and how to properly show all important sides of the image. For our dataset, we shot complete single product images with a white background under optimal lighting conditions, using the HAVOX HPB-40 Photo Studio. We made sure, every pictured side was visible from a straight angle with as little contortion as possible. This approach usually provided between two and six images, depending on the type of packaging and the shape of the product. Figure \ref{fig:ExampleImagesDataset} shows two examples of product images.

\section{\uppercase{Image Annotation}}

\label{sec:ImageAnnotation}

We classified and annotated all images and products. The following subsections detail our approach for selecting and specifying image and product annotations, as well as object detection labels. 

Note, that although we have worked very carefully, we acknowledge that inconsistencies may arise in our labelled data, due to varying interpretations of the specifications and human errors, despite multiple rounds of correction and annotation, potentially resulting in gaps and issues in the dataset that may go unnoticed.

\subsection{Data Specifications}
\label{sec:ImageAnnotation:subsec:DataSpecifications}

In this section, we provide an overview of our data structure, as well as which properties can be found in which file.

\begin{enumerate}
    \item Product Image [JPG]: All images were shot in JPG format, using a 1:1 aspect ratio, as described in the GS1 specification for creating product images \cite{noauthor_gs1_nodate}. We used the Canon EOS 2000D with a resolution of 4000x4000, to create the majority of product images, as well as the Canon EOS R with a resolution of 4480x4480 for a smaller subset of images. 
    \item Object Labels [XML]: Each image has its respective XML file, containing labels for objects found on the image. The labels are relevant for the training of object detection models. The object labels for each image are stored in the Pascal VOC Format \cite{everingham_pascal_2010}.
    \item Image Information [CSV]: The image information file for each image includes the following information:

    \begin{itemize}
        \item Image Type: Scope of how the product is shown in the image, according to the GS1 product image specification \cite{noauthor_gs1_nodate}.
        \item Facing: Face of the product in the image.
        \item Packaging Type: The type of packaging the product is in, as specified in \cite{netherlands_codes_nodate}. 
        \item Packaging Material: Material of the product’s packaging
        \item Fill Type: Number of single products in packaging (Single- / Multipack)
    \end{itemize}

    \item Product Information [CSV]: This file contains information about the properties of the products, like nutritional data, brand name, GTIN, or alcohol content. This correlates to the properties of a subset of object detection labels, central to identifying the product. Additionally, the file contains the GPC brick classification of the product. In total, the file has 30 attributes, 15 of which, belong to the nutrition table.
\end{enumerate}

\begin{figure}
    \centering
    \includegraphics[width=\linewidth]{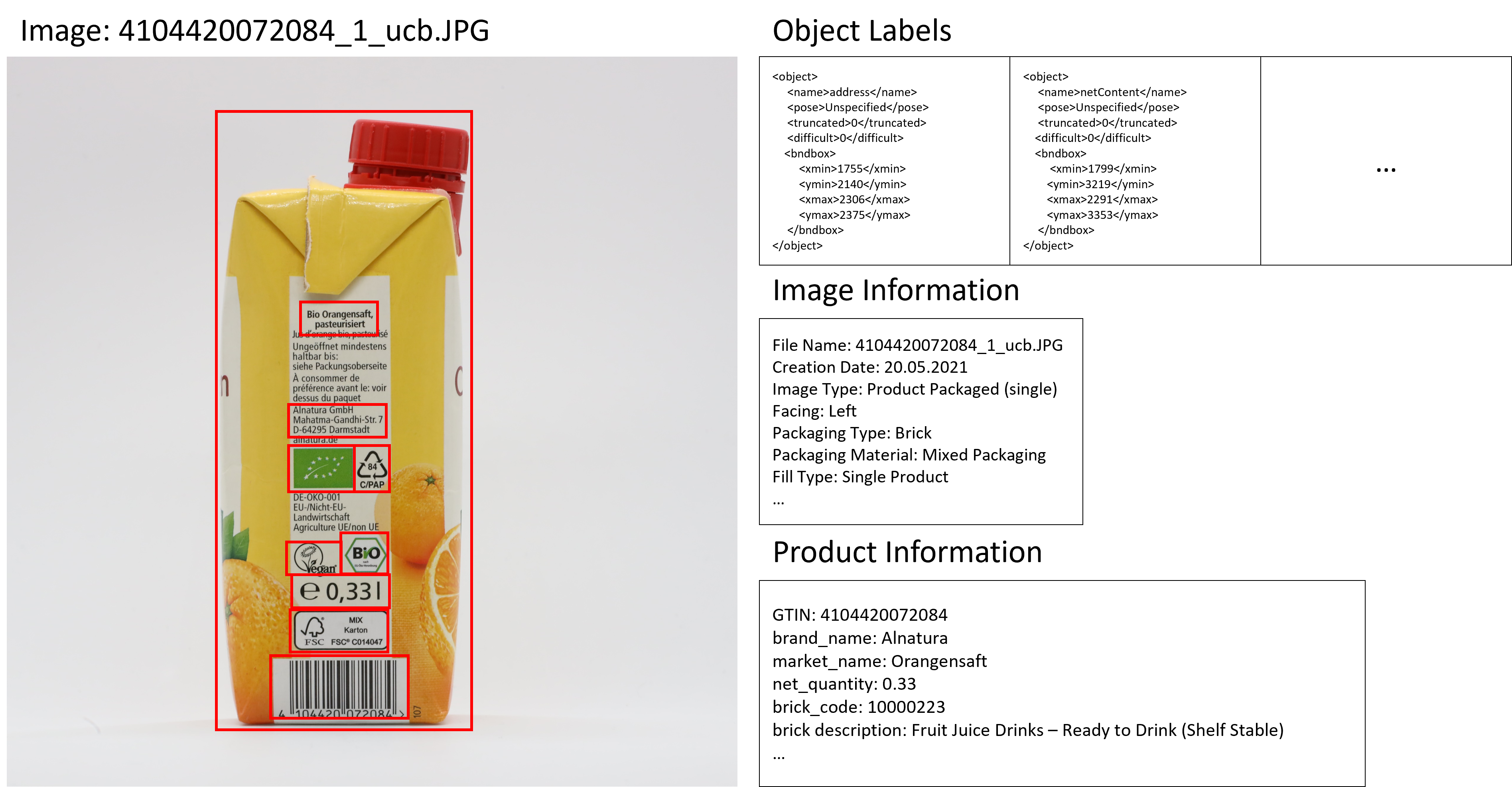}
    \caption{Example data for a single product image. Shows the product image with visualized object detection labels (left) and example data of label files for object detection labels (top right), image information (centre right), and product information (bottom right).}
    \label{sec:ImageAnnotation:subsec:DataSpecifications:fig:ExampleData}
\end{figure}

\subsection{Object Detection Labels}
\label{sec:ImageAnnotation:subsec:ClassesAndAnnotations}

\begin{figure*}[!ht]
    \centering
    \begin{subfigure}[t]{0.49\textwidth}
         \centering
         \includegraphics[width=\textwidth]{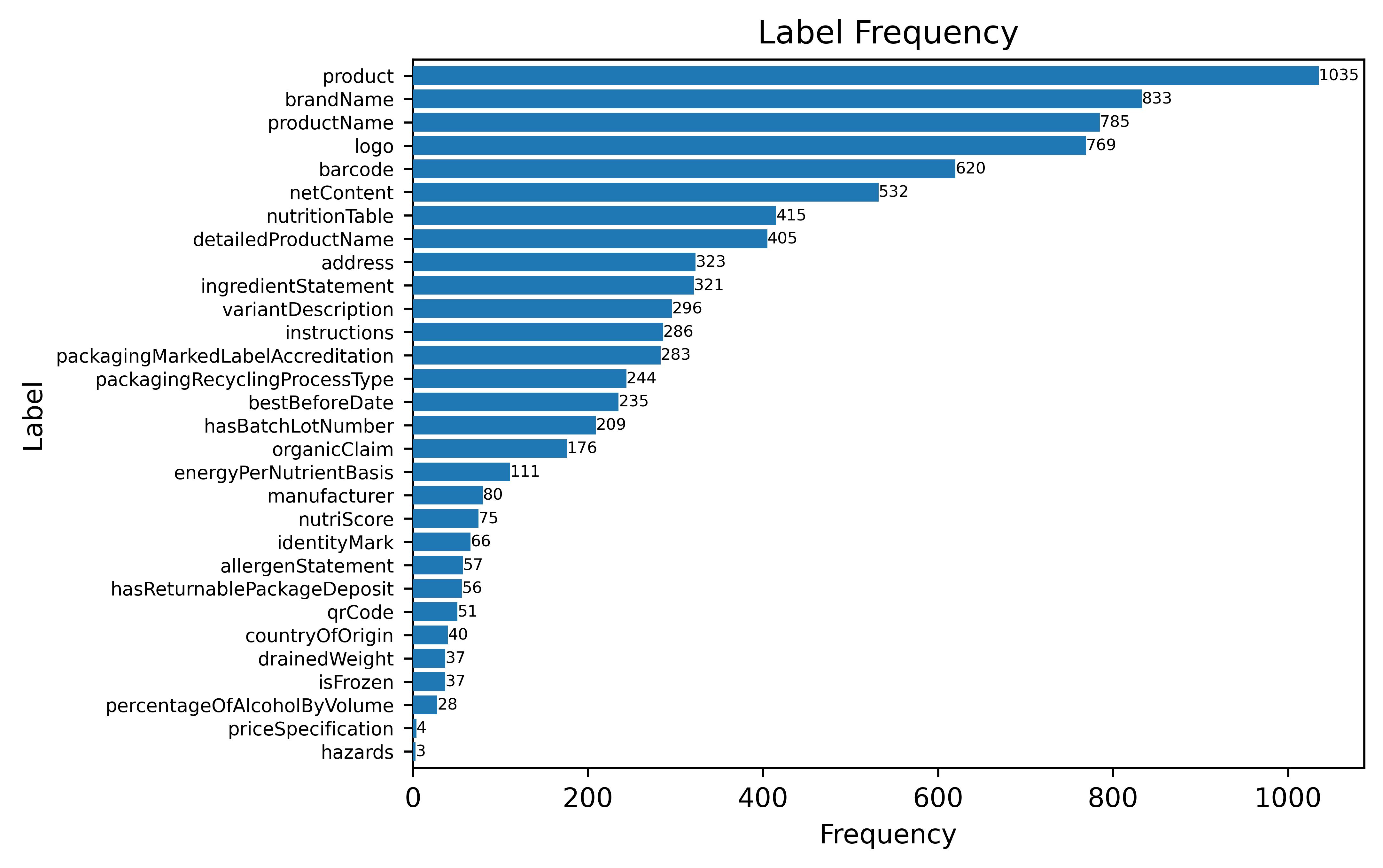}
         \caption{Distribution of all assigned labels.}
         \label{fig:Distribution}
    \end{subfigure}
    \hfill
    \begin{subfigure}[t]{0.49\textwidth}
        \centering
        \includegraphics[width=\textwidth]{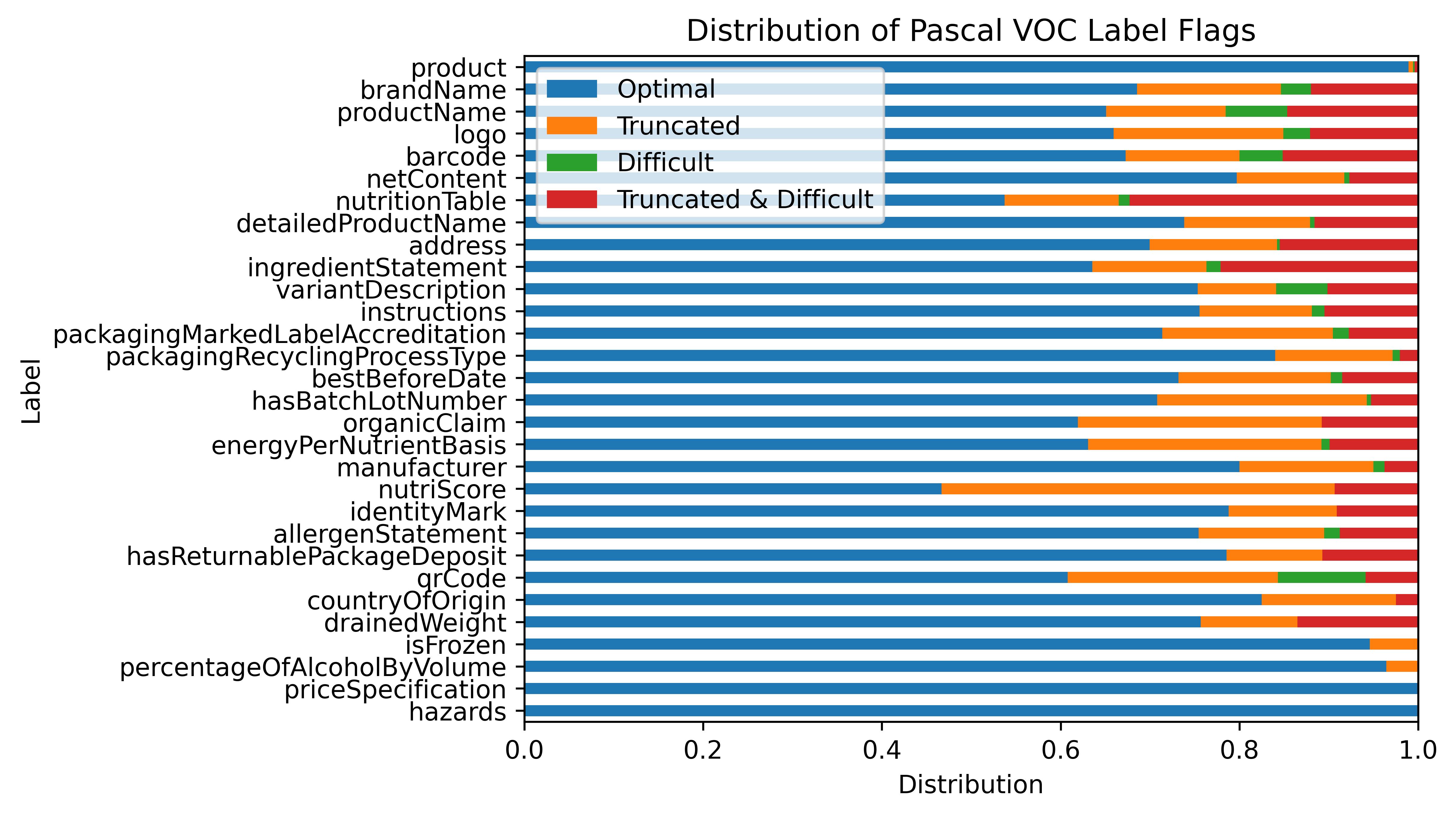}
        \caption{Distribution of the different combinations of label flags introduced by the Pascal VOC annotation format.}
        \label{fig:TruncatedDifficultDistribution}
    \end{subfigure}
    \caption{Statistical analysis of our dataset. N = 1034.}
    \label{fig:TwoGraphs}
\end{figure*}

The labels used for object detection were derived from common properties found on grocery packaging. For this, we first analysed, which elements are commonly found on grocery item packaging in Germany. Additionally, we took into account the regulations from the European Parliament to determine, which information is required on grocery item packaging (e.g. nutritional information, barcode, or information about alcohol content) \cite{european_parliament_and_of_the_council_regulation_2011}. Based on these commonly found packaging properties, we created a list of 30 labels that can be used to label different relevant areas on retail products, including one label marking the entire product. Where possible, we applied the naming convention provided by the GS1 Web Vocabulary \cite{noauthor_gs1foodbeveragetobaccoproduct_nodate}. A list of all labels can be found in figure \ref{fig:TwoGraphs} or table \ref{sec:BaselineModels:subsec:ObjectDetection:tab:Performance}.

\begin{figure}
    \centering
    \includegraphics[width=0.9\linewidth]{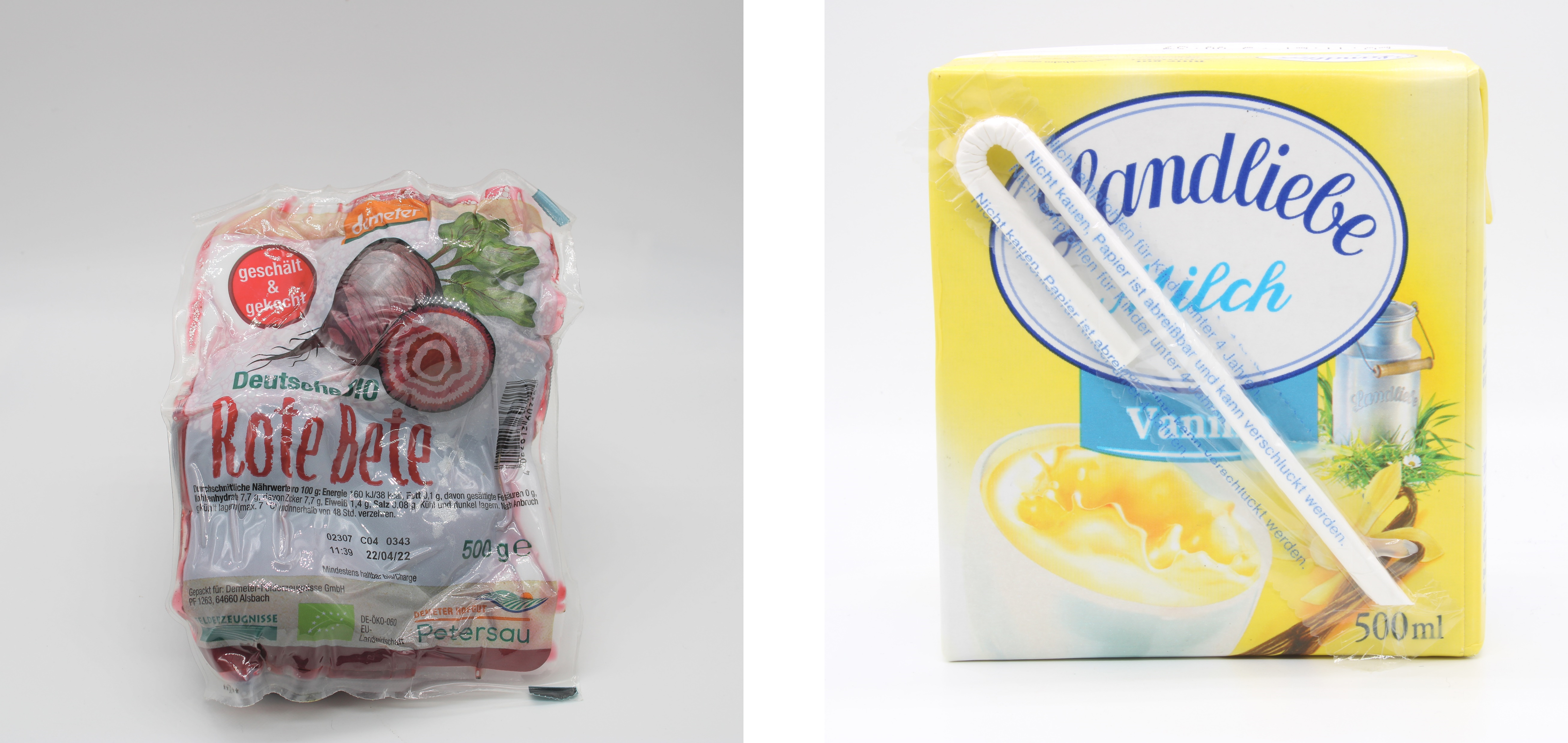}
    \caption{Example of images with difficult (left) and truncated (right) elements.}
    \label{fig:ExampleDifficultTruncated}
\end{figure}

We labelled every image, by drawing bounding boxes around every important area on the product and assigning the corresponding label to that area. In some cases, parts of the labelled object were cut off or difficult to read. Here, we assigned the optional truncated or difficult flag to the object. Whenever an object was only partially visible, due to it being cut off or part of the packaging overlapping it, we assigned the truncated flag to that label. In other cases, parts of a marked label were hardly readable (e.g. not in focus, blurred, or on a reflective surface). These labels were indicated with the difficult flag. Figure \ref{fig:ExampleDifficultTruncated} shows an example of a beetroot packaging that would be labelled difficult due to the reflections on the surface and milk packaging, where brand name and product name would be considered truncated due to the straw blocking some of the letters. The bounding boxes, including their label type and optional flags, were saved using the Pascal VOC format in an XML file. \cite{everingham_pascal_2010}. The XML file for each image has the same name as the image itself. For labelling, we used the tool LabelImg \cite{darrenl_labelimg_2022}, modified to allow for each label to include the difficult or truncated flags.

\subsection{Image Classification Labels}
\label{sec:ImageAnnotation:subsec:PropertryLabelling}
In addition to the assignment of object detection labels, we categorized each product and annotated the images. On the image level, we labelled, whether products were shown in or out of packaging, which side of the product is visible, how many single items are in one package, as well as types of packaging and material used to package the product. To differentiate between packaging containing a single or multiple items, we determined whether each item in the multipack is at least individually wrapped and contains a separate barcode. An example of this would be a six-pack of cans of beer. Packaging types were derived from the appropriate GS1 standard wherever possible \cite{netherlands_codes_nodate}. In cases, where no fitting packaging types was provided, we added our own names following the naming styles, used in the reference. The packaging sides shown in the image were split into categories front, left, back, and right. If the packaging was shown from a different point of view (e.g. top or bottom), the image was categorized as a "not-front" image. In some cases, it was impossible to tell, what face the image was taken from. In these cases, we labelled the side as "unspecified". We labelled all image properties using Label Studio \cite{noauthor_heartexlabslabel-studio_2023}. On product level, we assigned each product its corresponding GPC brick. This also allows for classification on class and family level.

\subsection{Extracted Property Values}
\label{sec:ImageAnnotation:subsec:PropertryLabelling:subsubsec:ProductPropertyValues}
In order to evaluate our future end-to-end system for product property extraction, we also extracted ground truth values for detected properties. This includes standard properties, such as product name, brand name, and GTIN, more specific properties, such as nutritional information, fill weight, or alcohol content, and which kinds of seals are found on the product. We differentiate between organic seals (such as the "bio" label), product and packaging quality seals (e.g. vegan, vegetarian) and labels showing whether the packaging of beverages could be returned to a vendor to be reused or not (called "Mehrweg" packaging in Germany). We only consider quality seals assigned by a third party. The most common seals were all categorized and indicated with numbers to be identified more easily. The resulting list contains seals found all across Germany, as well as more regional seals. 

\section{\uppercase{Statistical Description}}
\label{sec:StatisticalDescription}
Currently, the dataset contains a total of 250 products with 1,034 images, which averages 4.14 images per product. Most common are products with 4 images. 792 images were created using the Canon EOS 2000D, the other 242 images were shot with the Canon EOS R.

For object detection, we used a total of 30 labels. On average, there are 8.14 labels per image and about 33.65 labels across the images of a single product. Figure \ref{fig:Distribution} shows, how often a label occurs in the dataset. Labels found on multiple sides of a product, appear on nearly 80\% of all images, e.g. \textit{productName} or \textit{brandName}, while labels relevant to only one side of the product, such as \textit{nutritionTable}, appear more rarely. Some are also GPC family specific, like \textit{drainedWeight}, and therefore barely represented.

Figure \ref{fig:TruncatedDifficultDistribution} shows the distribution of difficult and truncated flags among the different labels. The distribution shows that a significant number of labels are either difficult, truncated, or both.

The most common facing types were front (273), back (242), and not-front (213). The most represented packaging types were cartons, boxes, and bags, with 97 products being packaged in plastic and 69 products packaged in paper.

A complete statistical description is provided by Jupyter Notebooks, which are included in the dataset repository.

\section{\uppercase{Baseline Models}}
\label{sec:BaselineModels}
To demonstrate the use of the dataset, we trained several deep-learning models. These models should be considered as baselines for their respective tasks. The models are general-purpose networks such as ResNet50 \cite{he_deep_2016} with very little adjustment to the problem presented. Therefore, we expect that these baseline models can be improved upon. All models have been chosen because of their popularity and availability as open-source software. Image classification results are shown in table \ref{sec:BaselineModels:fig:ClassificationResults}.

\begin{table}
    \caption{Performance of our baseline image classification models fine-tuned on a pre-trained ResNet50 \cite{he_deep_2016} implemented in the Torchvision package. Only the head of each model has been trained. In case of material and packagingType only classes with more than 50 instances were used for training. Metrics are weighted and averaged over all labels.}
    \begin{tabular}{|l|l|l|l|}
    \hline
              & material & packagingType    & facing \\ \hline
    Precision & 0.73     & 0.895            & 0.44   \\ \hline
    Recall    & 0.74     & 0.88             & 0.45   \\ \hline
    F1-Score  & 0.73     & 0.88             & 0.45  \\ \hline
    \end{tabular}
    \label{sec:BaselineModels:fig:ClassificationResults}
\end{table}

\begin{table*}[!ht]
    \small
    \caption{Results after training of the YOLOv5 model based on the default parameters \cite{jocher_yolov5_2020}. Results are briefly discussed in section \ref{sec:BaselineModels}. mAP@.5 and mAP@.5:.95 denote the mean average precision for different intersection over union thresholds as implemented in \cite{jocher_yolov5_2020}}.
    \centering
    \begin{tabular}{|l|l|l|l|l|l|l|}
    \hline
        class  & labels & precision & recall & mAP@.5 & mAP@.5:.95 \\ \hline
        all  & 1,665 & 0.698 & 0.496 & 0.521 & 0.394 \\ \hline
        address  & 69 & 0.450 & 0.246 & 0.299 & 0.189 \\ \hline
        percentageOfAlcoholByVolume  & 1 & 1.000 & 0.000 & 0.000 & 0.000 \\ \hline
        barcode  & 133 & 0.917 & 0.970 & 0.982 & 0.826 \\ \hline
        bestBeforeDate  & 40 & 0.748 & 0.700 & 0.712 & 0.450 \\ \hline
        brandName  & 166 & 0.538 & 0.536 & 0.541 & 0.352 \\ \hline
        energyPerNutrientBasis  & 19 & 0.945 & 0.737 & 0.771 & 0.630 \\ \hline
        countryOfOrigin  & 13 & 1.000 & 0.000 & 0.002 & 0.000 \\ \hline
        drainedWeight  & 11 & 0.884 & 0.545 & 0.642 & 0.536 \\ \hline
        variantDescription  & 44 & 0.385 & 0.318 & 0.255 & 0.132 \\ \hline
        isFrozen  & 7 & 0.282 & 0.121 & 0.049 & 0.040 \\ \hline
        hazards  & 1 & 1.000 & 0.000 & 0.000 & 0.000 \\ \hline
        identityMark  & 19 & 0.886 & 0.737 & 0.840 & 0.747 \\ \hline
        ingredientStatement & 62 & 0.512 & 0.387 & 0.371 & 0.225 \\ \hline
        instructions & 54 & 0.198 & 0.167 & 0.129 & 0.084 \\ \hline
        logo & 151 & 0.694 & 0.649 & 0.692 & 0.525 \\ \hline
        manufacturer & 19 & 0.392 & 0.263 & 0.319 & 0.248 \\ \hline
        productName & 140 & 0.572 & 0.621 & 0.608 & 0.344 \\ \hline
        detailedProductName & 85 & 0.455 & 0.259 & 0.259 & 0.155 \\ \hline
        nutriScore & 22 & 0.899 & 0.909 & 0.912 & 0.670 \\ \hline
        netContent & 90 & 0.784 & 0.789 & 0.787 & 0.614 \\ \hline
        nutritionTable & 87 & 0.821 & 0.828 & 0.831 & 0.702 \\ \hline
        organicClaim & 21 & 0.640 & 0.667 & 0.704 & 0.551 \\ \hline
        priceSpecification & 1 & 1.000 & 0.000 & 0.497 & 0.348 \\ \hline
        product & 213 & 0.877 & 0.995 & 0.995 & 0.954 \\ \hline
        qrCode & 19 & 0.961 & 0.737 & 0.747 & 0.556 \\ \hline
        packagingMarkedLabelAccreditation & 69 & 0.668 & 0.493 & 0.519 & 0.403 \\ \hline
        packagingRecyclingProcessType & 55 & 0.830 & 0.764 & 0.798 & 0.656 \\ \hline
        hasReturnablePackageDeposit & 9 & 0.878 & 1.000 & 0.995 & 0.657 \\ \hline
        hasBatchLotNumber & 37 & 0.389 & 0.324 & 0.315 & 0.195 \\ \hline
        allergenStatement & 8 & 0.323 & 0.125 & 0.046 & 0.029 \\ \hline
    \end{tabular}
    \label{sec:BaselineModels:subsec:ObjectDetection:tab:Performance}
\end{table*}
The object detection model is based on the YOLOv5 package\cite{jocher_yolov5_2020}. In our baseline training, we used the second-smallest variant, pretrained on the ImageNet dataset \cite{5206848}, with an image size of 1280x1280. We utilised an 80/20 split on product level, resulting in $200$ products ($821$ images) in the training- and $50$ products ($213$ images) in the validation set. Table \ref{sec:BaselineModels:subsec:ObjectDetection:tab:Performance} shows the results of our baseline training. We do not differentiate between labels which are marked as either truncated, difficult, or both, and those which aren't in our performance calculations. 
In the following, we list the key observations of our object detection approach according to the taxonomy of challenges in generic object detection as described in \cite{liu_deep_2020}:
\begin{itemize}
    \item High distinctness and low inner-class variation for labels \textit{barcode}, \textit{nutritionTable}, \textit{qrCode}, \textit{identityMark}, \textit{nutriScore} and \textit{netContent} lead to good results.
    \item High distinctness and high inner-class variation for labels \textit{brandName}, \textit{productName} and \textit{logo} lead to decent results. 
    \item Low distinctness and low inter-class variation of plain text labels, such as \textit{ingredientStatement} and \textit{detailedProductName} lead to unsatisfying initial results.
\end{itemize}

\section{\uppercase{Conclusions}}
\label{sec:ConclusionsAndFutureWork}

The dataset described in this paper consists of 250 different products with various annotations, labels and extracted properties. In our opinion, it represents an important first step and offers a valuable resource for researchers, who can use it to train and evaluate models for extracting product properties from product images. The globally available GS1 standards, as well as the provided documentation on labelling product images, allow users to extend the dataset beyond its current scope. Based on the specifications provided, extending the dataset involves the following steps:

\begin{itemize}
    \item Setting up the workstation: This takes approximately 5–10 minutes per photography session.
    \item Creating images: This process, which entails selecting the product, photographing it from all relevant angles, reviewing and renaming the images, takes around 5 minutes per product.
    \item Image annotation and labelling: This requires 20 minutes per product and involves entering product information into the appropriate CSV files and labelling the images.
\end{itemize}

The dataset facilitates evaluations of object detection, image classification and OCR models, while baseline models enable performance comparisons. 

As future work we plan to develop label-specific detection models, which are trained on the detection of a single label. In addition we plan to analyse the influence of the size of the training dataset on model performance. Based on our key observations listed in section \ref{sec:BaselineModels}, we assume that for labels with high inner-class variation and low distinctness (e.g. \textit{detailedProductName}, \textit{ingredientStatement}) more training data is necessary than for labels with low inner-class variation and high distinctness (e.g. \textit{barcode}, \textit{nutritionTable}) in order to improve the performance of the model. A first experiment on \textit{productName} and \textit{logo} with additional 3350 labelled front images was promising. Therefore, we plan to add more images with labels, annotations and property values to the described dataset.

Because extending the dataset requires a high amount of manual effort, we have developed a prototype for synthetic product image generation to increase the number of labelled product images. The generator creates 3D product objects, renders images, and assign label and bounding boxes based on rules, properties, and dependencies. For the sake of simplicity, we only looked at Tetra Packs in our initial implementation. Early results showed a lack of generalization from the YOLOv5 object detection model \cite{jocher_yolov5_2020} when trained on synthetic data. Therefore, future research will explore automating annotated product image creation using generative adversarial networks in combination with the developed rule based approach. We also plan to evaluate large language models such as GPT-4 \cite{openai_gpt-4_2023} as an alternative approach for extracting information from product images.

\section*{\uppercase{Acknowledgements}}
This work was funded by the German Federal Ministry of Education and Research (FKZ 01IS20085).

\section*{\uppercase{{Code Availability}}}
All Jupyter Notebooks, scripts, and data can be found inside the following repository: \url{https://gitlab.rlp.net/ISS/food-product-image-dataset}.

\bibliographystyle{apalike}
{\small
\bibliography{literature}}

\end{document}